%% file: main.tex
\documentclass[twoside]{article}

\usepackage[accepted]{aistats2025}
\usepackage[round]{natbib}

\input{math_commands}

\usepackage[utf8]{inputenc} 
\usepackage[T1]{fontenc}    
\usepackage[colorlinks,
            linkcolor=red,
            anchorcolor=blue,
            citecolor=blue,
            ]{hyperref}

\usepackage{hyperref}
\usepackage{url}
\usepackage{algorithm}
\usepackage{algpseudocode}
\usepackage{algorithmicx}
\usepackage{cleveref}
\usepackage{xcolor}
\usepackage{amssymb}
\usepackage{xspace}
\usepackage{graphicx}
\usepackage{pifont}

\newcommand{\alg}{{\texttt{DISC}}\xspace}
\newcommand{\ppv}{{\texttt{PPV}}\xspace}

\begin{document}

%
\runningtitle{Efficient and Asymptotically Unbiased Constrained Decoding for Large Language Models}

%
\runningauthor{Haotian Ye, Himanshu Jain, Chong You, Ananda Theertha Suresh, Haowei Lin, James Zou, Felix Yu}

\twocolumn[

\aistatstitle{Efficient and Asymptotically Unbiased\\Constrained Decoding for Large Language Models}

\aistatsauthor{ Haotian Ye$^{1*}$ \And Himanshu Jain$^2$ \And Chong You$^2$ \And Ananda Theertha Suresh$^2$}
\aistatsauthor{Haowei Lin$^3$ \And James Zou$^1$ \And Felix Yu$^2$}

\aistatsaddress{ $^1$Stanford University \And $^2$Google \And $^3$Peking University}


]


\begin{abstract}
  \input{abstract}
\end{abstract}

\input{Sections/introduction}
\input{Sections/preliminary}

\input{Sections/method}
\input{Sections/experiments}

\input{Sections/related_works}

\section{DISCUSSION \& CONCLUSION}
This paper focuses on comparing trie-based methods and does not consider combining \alg with countless existing techniques. It should be pointed out that \alg can be seamlessly integrated into existing language model frameworks without requiring any modifications to the underlying architecture. This minimally invasive property means it is compatible with most generation strategies such as temperature scaling or top-$k$ sampling, as well as task-specific techniques such as special tokenizations or retrieval tricks. 

In addition, notice that \alg and \ppv can be applied to other types of constraints (beyond set constraints $\mathcal S$ we consider). For instance, constraints defined by context-free grammars, specific formatting requirements like generating valid JSON structures, or other complex criteria can also be incorporated, and the theoretical analysis extends. Future work could explore empirical evaluations of \alg{} and \ppv{} across a broader spectrum of constrained decoding scenarios.

In conclusion, we address the challenges of bias and inefficiency in constrained decoding by introducing the \alg{} and the \ppv{}. We believe that our methods effectively mitigate unintended biases and enhance computational efficiency, broadening the applicability of constrained decoding in real-world scenarios.

\bibliography{reference}

\clearpage

\section*{Checklist}



 \begin{enumerate}

 \item For all models and algorithms presented, check if you include:
 \begin{enumerate}
   \item A clear description of the mathematical setting, assumptions, algorithm, and/or model. [Yes]
   \item An analysis of the properties and complexity (time, space, sample size) of any algorithm. [Yes]
   \item (Optional) Anonymized source code, with specification of all dependencies, including external libraries. [Yes]
 \end{enumerate}

 \item For any theoretical claim, check if you include:
 \begin{enumerate}
   \item Statements of the full set of assumptions of all theoretical results. [Yes]
   \item Complete proofs of all theoretical results. [Yes]
   \item Clear explanations of any assumptions. [Yes]     
 \end{enumerate}

 \item For all figures and tables that present empirical results, check if you include:
 \begin{enumerate}
   \item The code, data, and instructions needed to reproduce the main experimental results (either in the supplemental material or as a URL). [Yes]
   \item All the training details (e.g., data splits, hyperparameters, how they were chosen). [Not Applicable]
         \item A clear definition of the specific measure or statistics and error bars (e.g., with respect to the random seed after running experiments multiple times). [Not Applicable]
         \item A description of the computing infrastructure used. (e.g., type of GPUs, internal cluster, or cloud provider). [Yes]
 \end{enumerate}

 \item If you are using existing assets (e.g., code, data, models) or curating/releasing new assets, check if you include:
 \begin{enumerate}
   \item Citations of the creator If your work uses existing assets. [Yes]
   \item The license information of the assets, if applicable. [Not Applicable]
   \item New assets either in the supplemental material or as a URL, if applicable. [Not Applicable]
   \item Information about consent from data providers/curators. [Not Applicable]
   \item Discussion of sensible content if applicable, e.g., personally identifiable information or offensive content. [Not Applicable]
 \end{enumerate}

 \item If you used crowdsourcing or conducted research with human subjects, check if you include:
 \begin{enumerate}
   \item The full text of instructions given to participants and screenshots. [Not Applicable]
   \item Descriptions of potential participant risks, with links to Institutional Review Board (IRB) approvals if applicable. [Not Applicable]
   \item The estimated hourly wage paid to participants and the total amount spent on participant compensation. [Not Applicable]
 \end{enumerate}

 \end{enumerate}

\clearpage

\input{supplement}

\end{document}

%% file: math_commands.tex
\usepackage{amsmath,amsfonts,bm}
\usepackage{amsthm}

\newtheorem{theorem}{Theorem}[section]

\newtheorem{lemma}[theorem]{Lemma}

\usepackage{hhline}
\usepackage{diagbox}
\usepackage{arydshln}

\def\eqref#1{equation~\ref{#1}}

\def\1{\bm{1}}

\def\va{{\bm{a}}}

\def\vs{{\bm{s}}}
\def\vt{{\bm{t}}}

\def\vx{{\bm{x}}}
\def\vy{{\bm{y}}}

\DeclareMathAlphabet{\mathsfit}{\encodingdefault}{\sfdefault}{m}{sl}
\SetMathAlphabet{\mathsfit}{bold}{\encodingdefault}{\sfdefault}{bx}{n}

\usepackage{color}
\usepackage{multirow}
\usepackage{booktabs}

\usepackage{listings}
\usepackage{color}

\definecolor{codegray}{rgb}{0.5,0.5,0.5}
\definecolor{codepurple}{rgb}{0.58,0,0.82}

\lstdefinestyle{mypython}{
    language=Python,
    backgroundcolor=\color{white},
    commentstyle=\color{codegray},
    keywordstyle=\color{blue},
    stringstyle=\color{codepurple},
    basicstyle=\ttfamily\footnotesize,
    breaklines=true,
    frame=single,
    showstringspaces=false,
    captionpos=b,
}

%% file: abstract.tex
In real-world applications of large language models, outputs are often required to be confined: selecting items from predefined product or document sets, generating phrases that comply with safety standards, or conforming to specialized formatting styles.
To control the generation, constrained decoding has been widely adopted. 
However, existing prefix-tree-based constrained decoding is inefficient under GPU-based model inference paradigms, and it introduces unintended biases into the output distribution. 
This paper introduces Dynamic Importance Sampling for Constrained Decoding (\alg{}) with GPU-based Parallel Prefix-Verification (\ppv), a novel algorithm that leverages dynamic importance sampling to achieve theoretically guaranteed asymptotic unbiasedness and overcomes the inefficiency of prefix-tree. Extensive experiments demonstrate the superiority of our method over existing methods in both efficiency and output quality. These results highlight the potential of our methods to improve constrained generation in applications where adherence to specific constraints is essential.


%% file: Sections/introduction.tex
\section{INTRODUCTION}

Large language models (LLMs) \citep{lee2023benefits,gilbert2023large,hwang2023review, skjuve2021my} exhibit remarkable capabilities in a variety of tasks and have been widely used in different applications, including question answering \citep{brown2020language,black2022gpt}, coding \citep{achiam2023gpt}, vision understanding \citep{liu2023visual,zhu2023minigpt4,ye2023mplugowl}, mathematics \citep{trinh2024solving}, and more. 
Conventional autoregressive decoding does not restrict the tokens generated in each decoding step, which is reasonable in general use cases.  
However, many real-world scenarios demand particular control over the output of these models. 
For instance, the user may require the outputs to belong to (or be excluded from) a huge predetermined set \citep{Cao2021,rajput2024recommender}, adhere to specific formats \citep{Geng2023}, or conform to particular grammars such as context-free grammars \citep{Kellner2024}.

Recent studies have demonstrated \textit{constrained decoding} as a helpful technique for aligning the output of LLM with requirements \citep{sun2024learning,Cao2021}. 
Given a predefined set of allowed outputs, it dynamically restricts the permissible tokens in each decoding step such that the final output aligns with the predefined constraints, enabling numerous applications where adherence to user-defined requirements is critical.
For example, in recommendation systems where the model is asked to generate an identifier for a product code, URL, or legal term from a large database, constrained decoding ensures that the generated text corresponds to exactly one of the valid identifiers \citep{bevilacqua2022autoregressive}. Similarly, in content filtering, constrained decoding can prevent the generation of prohibited words or phrases by restricting to a set of valid tokens \citep{zhang2023safetybench}.


\begin{figure}[h]
    \centering
    \includegraphics[width=\linewidth]
    {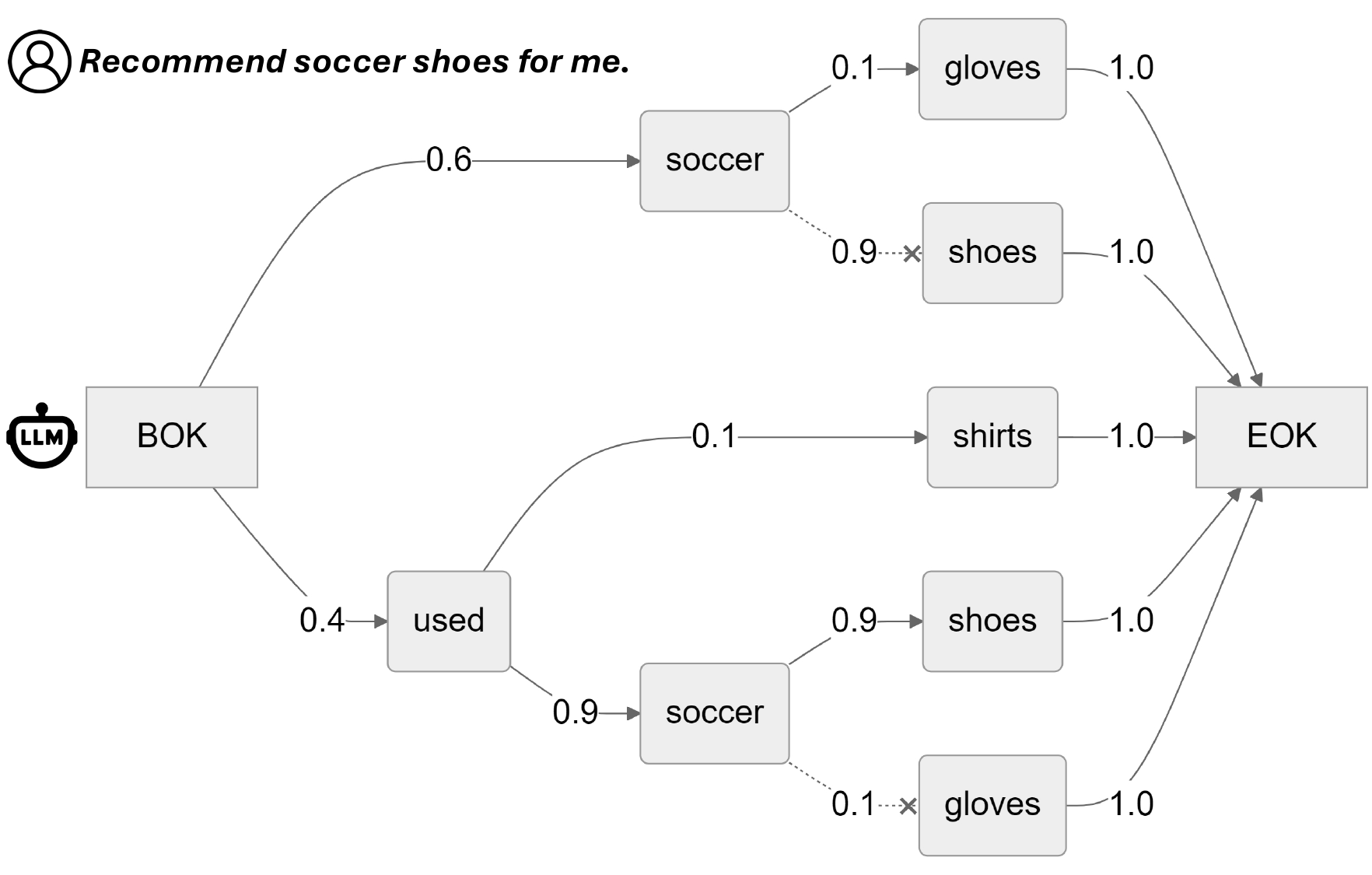}
    \vspace{-10pt}
    \caption{
    An illustration of how biased generation occurs when constrained decoding is applied. A user requests the model to recommend a product related to ``soccer shoes''. The probability of selecting each token is represented by the connection lines, where dashed lines with \ding{53} indicate invalid tokens (e.g., ``soccer shoes'' and ``used soccer gloves'' are not available in the recommendation list). Even though the probability of ``soccer gloves'' ($0.06$) is much lower than ``used soccer shoes'' ($0.324$), the model generates the former with a higher final probability ($0.6 \times 1$) than the latter ($0.4 \times 0.9 \times 1$), since the model shortsightedly selects ``soccer'', unaware of the invalidity of ``soccer shoes''.
    }
    \label{fig:bias_illustration}
\end{figure}


While ensuring compliance with user requirements, we demonstrate in this paper that constrained decoding introduces unintended biases into the output distribution, a critical issue that existing works have not addressed \citep{sun2024learning}. 
As illustrated in \Cref{fig:bias_illustration}, this bias stems from the unawareness of future constraints during autoregressive decoding. 
The model may start with generating high-probability tokens that could ultimately lead to sequences that violate the constraints. 
Forced by constrained decoding to adjust, the model has to select subsequent tokens with lower probability, resulting in outputs that do not truly represent the model's learned language patterns. 
In \Cref{thm:lower_bound}, we quantify the universal impact of bias across diverse models, tasks, and datasets. 
This significantly degrades the quality and reliability of the generated outputs, posing severe challenges for applications that rely on accurate and unbiased generation.

Moreover, existing constrained decoding methods use prefix trees (a.k.a., \emph{tries}), a data structure used in file systems \citep{bayer1977prefix}, to store all valid outputs. 
However, these tries are typically implemented on CPUs, leading to a heavy data communication overhead when integrating them with language models running on GPUs \citep{owens2008gpu} or TPUs \citep{jouppi2017datacenter}. 
Each time a token is generated, it must be transferred to CPUs to query the trie, incurring substantial communication overhead.
While implementing tries on GPUs can reduce the communication cost, their fragmented structures and varying node sizes pose significant challenges for efficient parallel processing. 
These inefficiencies hinder the practical deployment of constrained decoding in real-world applications where speed and scalability are critical.


To address these challenges, we propose a novel approach that is asymptotically unbiased while enhancing computational efficiency for set-based constrained decoding. Our method comprises two key contributions: a \textbf{Dynamic Importance Sampling algorithm for Constrained decoding} (\alg), equipped with a \textbf{Parallel Prefix-Verification} (\ppv) subroutine optimized for high-performance computing environments. \alg dynamically extends existing importance sampling techniques to LLMs decoding and provides a theoretically guaranteed sampling method that adjusts the token probabilities during decoding to reflect the true distributions under the constraints more accurately. As proved in \Cref{thm:alg}, \alg is unbiased with a fixed expected sampling size, ensuring that the generated outputs are both compliant with the constraints and statistically consistent with the language model's learned distribution.

Regarding the efficiency issue, \ppv significantly accelerates the decoding by alternatively ``verifying'' rather than ``returning'' all valid candidate tokens. By leveraging the fact that the language model already computes the scores for all tokens at each decoding step, it verifies only tokens with top scores to determine whether they are valid under the given constraints. 
This removes the requirements on prefix-tree and can be realized with a ordered candidate array with logarithmic time complexity $\mathcal O(\log N)$, where $N$ is the total number of candidates in the constraint set. 
Importantly, the entire verification process is therefore highly parallelizable across all batches and top tokens. This bypass the inefficient trie loading, frequent data transfers between devices, and fragmented data access, significantly reducing computation time.

We run on 20 datasets across four tasks to evaluate our proposed method compared to existing baselines. This includes existing benchmarks such as entity disambiguation and document retrieval, as well as novel use cases such as generative retrieval under in-context learning.
Not surprisingly, we consistently outperform trie-based methods across all datasets while being at most $8.5$ times faster. Remarkably, our method can be understood as a meta-algorithm that can be applied to different techniques for certain applications, as it only improves the constrained decoding process.

We summarize our contributions as follows:
(1) We uncover and theoretically analyze the bias introduced by constrained decoding in autoregressive models, highlighting its universal impact on output distributions.
(2) We propose \alg equipped with \ppv, a novel algorithm that provides asymptotically unbiased sampling for constrained decoding with impressive efficiency. Remarkably, it can be seamlessly integrated with other generation strategies and adapted to various constraints.
(3) We provide comprehensive experimental results across various datasets and tasks, validating the effectiveness and efficiency of our approach.
In summary, we aim to enhance the reliability and fairness of LLM-generated content, thereby broadening the applicability of these powerful models. \footnote{Code is available at \url{https://github.com/YWolfeee/Large-Scale-Selection-for-LLMs}.}

%% file: Sections/preliminary.tex
\section{PROBLEM DEFINITION}

\input{Sections/alg}

This section introduces notations for language modeling and background on trie-based constrained decoding. 
We also explain the two major challenges of existing approaches, namely the biasedness of constrained sampling and the inefficiency of trie implementation. 

\subsection{Notations and Background}
Throughout the paper, we use $\vx = [x_1, x_2, ..., x_T]$ to represent a sequence where $|\vx| = T$ represents the sequence length.
We use $\vx_{<t} \triangleq [x_1, \cdots, x_{t-1}]$ to represent the sub-sequence of $\vx$ containing the first $t-1$ entries. 
We say that a sequence $\vy$ is a prefix of the sequence $\vx$, denoted as $\vy \preceq \vx$, if there exists an integer $t \le T$ such that $\vy = \vx_{<t}$.
Given two sequences $\vx$ and $\vy$, we use $\vx + \vy$ to represent their concatenation. 


\textbf{Autoregressive language modeling} is the backbone of contemporary LLMs that outputs text in a \textit{token-by-token} manner,  closely mimicking the generation process of natural languages. 
Mathematically, let $\mathcal V$ be a pre-defined vocabulary (i.e., token) set\footnote{Mathematically, we assume that $\mathcal V$ is equipped with an order. This holds trivially in modern computers when tokens are represented as integers.}, a sentence is represented as a sequence $\vx = [x_1,\cdots,x_T]$ where $x_i \in \mathcal V, i = 1, \ldots, T$ are individual tokens. 
The autoregressive language modeling factorizes the probability of generating $\vx$ as 
\begin{align*}
    P(\vx) = \prod_{t=1}^T P(x_t|\vx_{<t}).
\end{align*}
Then, each conditional probability $P(x_t|\vx_{<t})$ may be estimated from the output of a language model $L$, which we represent as $P_L(x_t| \vx_{<t})$. 
Notice that here we use $L$ to denote the language model with input prompts equipped, i.e., $P_L(x)$ is the output distribution conditioned on inputs that are part of $L$, since our study is regardless of the inputs that users provide.
We will omit the subscript $L$ for simplicity if there is no ambiguity.

\textbf{Constrained decoding} is a technique to align the output of a language model with particular constraints. 
For this paper, we focus on \textit{set constraints}, i.e., the generated sequence is required to be from a pre-defined set $\mathcal S$ of sequences, which in practice could be a particular set of products, paragraphs, documents, etc.
In autoregressive language generating, this requires that $P(x_t|\vx_{<t})$ is zero for all $x_t \in \mathcal V \setminus \mathcal A_{\vx_{<t}, \mathcal S}$ where $\mathcal A_{\vx, \mathcal S}$ for general $\vx, \mathcal S$ is a \emph{valid set} defined as
\begin{align*}
    \mathcal A_{\vx, \mathcal S} \triangleq \{v \in \mathcal V : \exists \vs \in \mathcal S ~~\text{s.t.}~ \vx + [v] \preceq \vs\},
\end{align*}
or unless the model generates the \textit{end-of-keyword} (\texttt{EOK}) token when $\vx_{<t} \in \mathcal S$. 
A typical way of checking if a token is in the valid set is through encoding $\mathcal S$ with a prefix-tree, which we explain next. 

\textbf{Prefix-tree}, also known as trie, is a type of search tree for checking if a particular string is from a given set of strings. 
Each node in the tree represents a token in $\mathcal V$ with the root node representing the \textit{begining-of-keyword} (\texttt{BOK}) token.
Given a path from the root to node $\vx$, a node $V \in \mathcal V$ is the child of the node if and only if $V \in \mathcal A_{\vx, \mathcal S}$. 
When generating a sequence using a language model, the algorithm starts from the root node and keeps generating a token that is the child of the last generated token until the entire sequence is returned (reached a leaf node). As an illustration, in \Cref{fig:bias_illustration}, only ``soccer gloves'', ``used shirts'', and ``used soccer shoes'' can be generated. 
We denote the output distribution of constrained decoding as $P^\text{CD}_\mathcal S$.

\subsection{Bias Induced by Constrained Decoding}
Ideally, given a language model $L$, we want to sample from the distribution $ P _\mathcal S (\vx) = P_L(\vx) / P_L(\mathcal S)$. Naively, one can score all sequences in $\mathcal S$ and sample in hindsight, but this is impractical when $\mathcal S$ is large, a typical setting in constrained decoding.
Although prefix-tree-based constrained decoding offers an alternative to sample, it inevitably induces bias in output distribution, a problem that existing works fail to consider. 

Intuitively, the bias can be attributed to the myopic autoregressive decoding when additional constraints are imposed: the model can start generating a prefix $\vx \preceq \vy$ when $P_L(\vy)$ is large, but ends up generating a low-probability sequence $\vy'$ where $\vx \preceq \vy'$, because $\vy \notin \mathcal S$ and $\vy' \in \mathcal S$. Remarkably, this is an issue only when constrained decoding is applied, because in conventional model generation, suffices with high probability will not be masked out. Below, we present in theory how biased the output is depends on the ``bad'' probability $p_b$ that the model assigns on sequences outside $\mathcal S$, i.e., $1 - P_L(\mathcal S)$.


\begin{theorem}
\label{thm:lower_bound}
        For arbitrary probability value $p_b$, there exists an autoregressive model $L$ and a candidate set $\mathcal S$ with $p_b = 1 - P_L(\mathcal S) $, such that the KL divergence\footnote{For two discrete distribution $P, Q$ over set $\mathcal X$, the KL divergence is defined as $KL(P\|Q) = \sum_{x\in \mathcal X}P(x)\ln \frac{P(x)}{Q(x)}$.} between $ P_{\mathcal S}$ and $ P^{\text{CD}}_\mathcal S$ is lower bounded by 
        \begin{align}
            KL( P_{\mathcal S} \|  P^{\text{CD}}_\mathcal S) \geq \Omega \Big( \ln\frac{1}{1 - p_b} \Big).
        \end{align}
\end{theorem}

The proof of \Cref{thm:lower_bound} is deferred to the appendix. It implies that the bias is universal, regardless of language models and tasks. So long as the probability outside $\mathcal S$ is large, the divergence can be significant.



\subsection{Inefficiency of Constrained Decoding}

In addition to the estimation error, trie-based constrained decoding methods also suffer from efficiency issues. 
Theoretically, the time complexity for constrained decoding is $\mathcal O(|\vx||\mathcal V|)$, and for large $\mathcal S$ with large size, $|\vx|$ is typically in the order of $\log(|\mathcal S|)$.
In comparison to the LLM inference time, this might seems negligible.
However, since LLMs are run on advanced computation devices such as GPUs and TPUs, having a trie run on CPUs means \textit{at each decoding step}, current prefixes have to be sent to CPUs for constrained decoding, and all valid subsequent tokens have to be sent back. Even constrained decoding can be run in parallel in CPUs, data communication itself is highly inefficient, incurring substantial communication overhead in integration with high-throughput GPU-based LLM inference.

A seemingly simple solution is to have a trie ran purely on GPUs for constrained decoding. However, to the best of our knowledge, whether an efficient GPU trie algorithm exists remains to be studied in the first place. 
Two fundamental challengs have to be addressed.
First, as the number of children can vary substantially from node to node, maintaining a tire that consists of a huge number of fragments with different sizes on GPUs, even if not impossible, is difficult and unfriendly to parallel computation. 
Second, it is unclear how to search multiple candidates on tries in a vectorized manner, such that the resources of GPUs can be fully leveraged. Given that the stopping time and size of children node are intrinsically different from candidates to candidates, organizing them into vectors for parallel computation remains difficult.

Overall, conventional trie-based constrained decoding is unsuitable with LLMs in terms of estimation accuracy and efficiency. Remarkably, this misalignment is irrespective of how the constraints are represented (e.g., whether keywords are sentences, phrases, or ID lists) and how LLMs are capable. It is rather a fundamental issue impeding their downstream applications.

%% file: Sections/alg.tex
\begin{algorithm*}[t] 
\caption{\textbf{D}ynamic \textbf{I}mportance \textbf{S}ampling for \textbf{C}onstrained Decoding (\alg{})
} 
\label{alg:disc} 
\begin{algorithmic}[1] 

    \State \textbf{Input:} Language model $L$, constraint set $\mathcal S$, maximum allowed sampling steps $K$, temperature $T$ 

    \Function{SampleCandidate()}{}:
        \State Initialize candidate sequence $\va = []$, importance score $x = 1$.
        \While{$\va \notin \mathcal S$}
        \State Get next token distribution $P = L(\va, T)$.
        \State Get mask of valid tokens $\texttt{mask} = \texttt{PPV}_{\mathcal{S}}(P, \va)$. \label{code:line_ppv}\Comment{Call \Cref{alg:ppv}}
        \State Sample a token $t$ from the distribution $\frac{P \,\odot\, \texttt{mask}}{\left| P \,\odot\, \texttt{mask} \right|_1}$.
        \State Update $\va = \va + [t]$, $x = x \times \left| P \odot \texttt{mask} \right|_1$.
        \EndWhile
        \State \Return $\va$, $x$
    \EndFunction
    \State Initialize sampling step $k = 0$ 
    \While{$k < K$}     \Comment{Sample at most $K$ candidates}
        \State $(\va, x) = \texttt{SampleCandidate}()$
        \If{$\va \in \mathcal{S}$}
            \State Update $k = k + 1$.
            \If{$x > \epsilon \sim \mathcal{U}[0,1]$} \Comment{Accept with probability $x$}
                \State \textbf{Return:} Sampled candidate sequence $\va$.
            \EndIf
        \EndIf
    \EndWhile

    \State Sample $(\va_k, x_k) = \texttt{SampleCandidate}()$ for $K$ times, then sample $\va$ from $\{\va_k\}_{k=1}^K$ with probability $\frac{x_k}{\sum x_k}$. \label{code:IS}
    \State \textbf{Return:} Sampled candidate sequence $\va$ 
\end{algorithmic}
\end{algorithm*}

%% file: Sections/method.tex
\section{METHOD}



This section presents our method to mitigate the bias and improve computational efficiency. 

\subsection{Dynamic Importance Sampling for Constrained Decoding (\alg{})}

\alg incorporates importance sampling into LLM decoding.  
Unlike traditional fixed-size importance sampling, \alg dynamically adjusts the sampling process, resulting in better efficiency and convergence rate.

The pseudo-code of \alg{} is in Algorithm \ref{alg:disc}.
It keeps sampling candidate sequences using standard constrained decoding and accepts each sample with a probability computed from how well the unconstrained output aligns with the constraints. 
This process is continued until a candidate sequence is accepted, upon which the distribution is guaranteed to be unbiased. 
If all $K$ samples are rejected, we further run an importance sampling on $K$ new sample sequences to minimize the bias.
Notice that the algorithm calls a core function $\texttt{PPV}(P, a)$ that returns a mask of valid tokens based on the current partial sequence $a$ and the token probability distribution $P$. This sub-routine is GPU-optimized and is presented in the following subsection.


\paragraph{Theoretical Analysis.}
The following theorem states that \alg{} achieves an upper bound on the Kullback-Leibler (KL) divergence between the distribution induced by \alg and the true constrained distribution $P_{\mathcal{S}}$. 

\begin{theorem} 
\label{thm:alg} 
For any autoregressive model $L$ (equipped with any text prompt), denote $P^{K}_{\text{IS}}$ the distribution induced by \Cref{alg:disc} with a maximum of $K$ sampling steps under constrained decoding. Then, the KL divergence between $P^{K}_{\text{IS}}$ and the true distribution $P_{\mathcal{S}}$, $\text{KL}\left(P_{\mathcal{S}} \| P^{K}_{\text{IS}}\right)$ is bounded by 
\begin{align} 
\mathcal{O}\left( 
\frac{p^K_b}{1 - p_b} \cdot \Big( 
\sqrt{\frac{p_b(1-p_b)}{K}}+ \frac{p_b}{K}
\Big)
\right), 
\end{align} 
where $p_b = 1 - P_L(\mathcal{S})$ is the total probability mass of sequences outside $\mathcal S$, and $\mathcal{O}$ hides universal constants. In addition, the expected number of sampling steps is
\begin{align*}
    \mathbb E [k] = \frac{1 - p_b^K}{1 - p_b} + K p_b^K  \leq \frac {1+e}{1 - p_b}.
\end{align*}
Furthermore, the algorithm is unbiased, i.e., 
\begin{align*}
    KL(P_\mathcal S \| P_{IS}^K) \to 0,
\end{align*}
with finite expectation number $\mathbb E[k] \to \frac{1+e}{1 - p_b}$.
Here the expectation is over all LLM sampling, acceptance/rejection procedure, and importance sampling. 
\end{theorem}

The proof of \Cref{thm:alg} is provided in the appendix. 
It shows that the estimator is asymptotically unbiased: when $K \to \infty$, the bound shrinks to zero, i.e., the estimated distribution converges to the ground-truth, and the convergence is \textit{exponentially} fast. 
Even if we set $K = \infty$ so that the divergence is zero, the expected sample step is still bounded, implying the unbiased constrained decoding is realizable.
Otherwise, the expected number of sampling steps is still generally much smaller than $K$ when $p_b$ is not close to $1$. 
Finally, if the invalid probability $p_b$ goes to zero, the unconstrained decoding almost generates valid sequences, and the bound shrinks to zero. 
This is expected because the constrained decoding does not (need to) take any effect.
One may fine-tune the language model on $\mathcal S$ to reduce the bound in practice. 




\subsection{Parallel Prefix-Verification (\ppv)}

\begin{figure}
    \centering
    \includegraphics[width=0.98\linewidth]{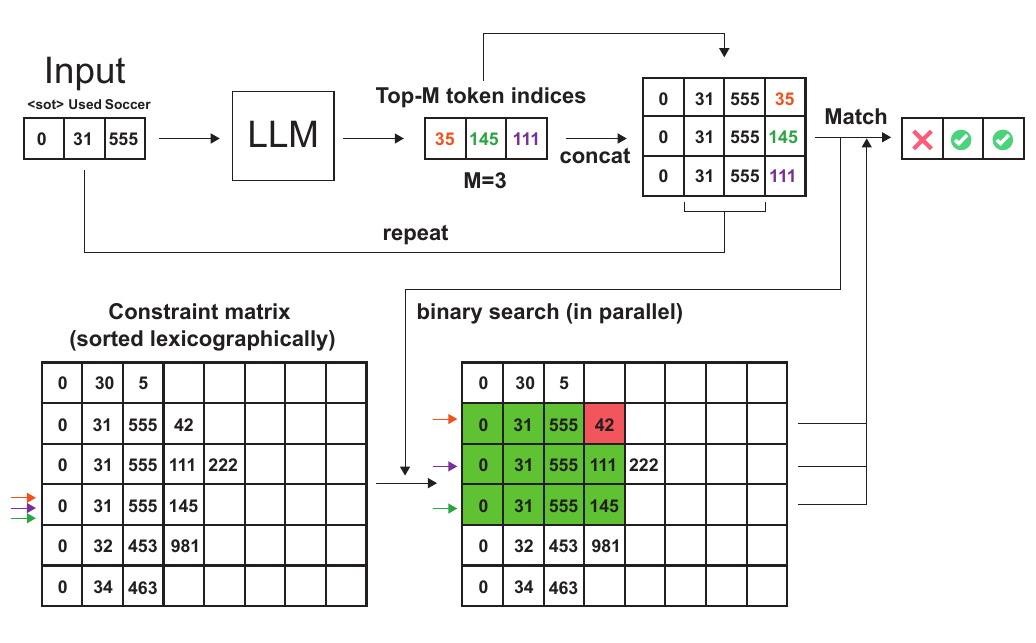}
    \caption{An illustration of \ppv. Assume the input is $[0,31,555]$, and the top-three token candidates given by the LLM are $\{35, 145,111\}$. \ppv concatenates them to the input and forms a three-row matrix. Then, it compares each row with the prepared array $\mathcal X$ where each row is a keyword in $\mathcal S$, in parallel. Since $\mathcal X$ is ordered alphabetically, the comparison can be performed via a binary search. After the comparison, we can determine whether each of the three partial outputs is the prefix of a row in $\mathcal X$. \ppv finishes by returning a vector mask indicating the validity of each candidate.
    }
    \label{fig:ppv_illustration}
\end{figure}

We next turn to the efficiency issue.
The core operation in constrained decoding is to mask out invalid tokens given already generated prefixes, as specified in \Cref{code:line_ppv} of \Cref{alg:disc}.
We develop the Parallel Prefix-Verification (\ppv) subroutine that can benefit from GPU parallel processing by verifying whether the tokens with the highest probabilities from the language model are valid in each decoding step, avoiding complicated data transfers and trie traverse.
We provide the pseudo-code in \Cref{alg:ppv} with details below, and present an illustration in \Cref{fig:ppv_illustration}.

\begin{algorithm}[ht]
\caption{\textbf{P}arallel \textbf{P}refix-\textbf{V}erification (PPV)} \label{alg:ppv} 
\begin{algorithmic}[1] 
    \State \textbf{Prepare:} Represent the constraint set $\mathcal{S}$ by a matrix $\mathcal X \in \mathcal V^{|\mathcal S| \times l_\text{max}}$ with rows sorted lexicographically based on the vocabulary order on $\mathcal V$. Pad shorter sequences to $l_\text{max}$, i.e., the maximum length of sequences in $\mathcal S$.\label{code:prepare}
    \State \textbf{Input:} Token distribution $P$, partial sequence $\va$
    \State Compute the top $M$ entries of $P$ and record their indices $\vt = \{t_1, t_2, \dots, t_M\}$. \label{code:topk}
    \State Construct a candidate matrix $V \in \mathcal{V}^{M \times (|\va| + 1)}$ where each row $V_i = \va + [t_i]$. \label{code:vector}
    \State Simultaneously perform a binary search for each candidate sequence $V_i$ in the rows of $\mathcal X$ to determine if $V_i$ is a prefix of any sequence in $\mathcal{S}$. Record the results in a boolean array $\texttt{mask}_V \in \{0,1\}^M$:
        \[
        \texttt{mask}_V[i] = 
            \begin{cases} 
                1 & \text{if } \exists j  ~\text{s.t.}~  V_i \preceq \mathcal X _j  \\
                0 & \text{otherwise}
            \end{cases}
        \]
    \State Extend $\texttt{mask}_V$ to $\texttt{mask}  \in \{0,1\}^{|\mathcal V|} $ with $\texttt{mask} [t_i] =1$ if $\texttt{mask}_V[i] =1$ and the remaining to $0$.
    \State \textbf{Return:} Mask of valid tokens $\texttt{mask} $ 
\end{algorithmic} 
\end{algorithm}

\begin{table*}[t]
\centering
\caption{\textbf{Task Statistics}. ``\#Corpus'' refers to the number of valid entities, while ``Avg. corpus token length’’ represents the average number of tokens per entity. ``\#Query'' indicates the number of test queries for each task. ``Vocab size’’ denotes the vocabulary size of the generator LLM, and ``Metrics’’ specifies the evaluation metrics.}
\resizebox{\textwidth}{!}{\begin{tabular}{cccccc}
\toprule
\textbf{Task}         & \textbf{\#Corpus} & \textbf{Avg. corpus token length} & \textbf{\#Query} & \textbf{Vocab size} & \textbf{Metrics}  \\
\midrule
Entity Disambiguation & 5,903,530         & 7.701                             & 24,100            & 50,264              & Micro F1          \\
Document Retrieval    & 5,903,530         & 7.701                             & 51,304            & 50,264              & R-Precision       \\
Entity Retrieval      & 4,635,922         & 9.485                             & 400              & 128,256             & Relevance score   \\
Fact Checking         & 5,416,593         & 9.436                             & 1,535             & 128,256             & Relevance score  \\
\bottomrule
\end{tabular}}
\end{table*}

\paragraph{Array Preparation.} 
Given the constraint set $\mathcal{S}$, \ppv first performs a preprocessing in \Cref{code:prepare} so that it is amenable to binary search during the subsequent verification step. 
Specifically, the sequences in $\mathcal{S}$ are ordered by first comparing their initial tokens according to the order specified by the vocabulary set $\mathcal{V}$. If the initial tokens are identical, the comparison proceeds to the subsequent tokens in a left-to-right manner. A padding token is appended to the end of shorter keywords as necessary to ensure uniform sequence length. 
All ordered sequences are aggregated into an array $\mathcal X$ on GPUs, which can be efficiently binary searched during decoding. 
In practice, we also split $\mathcal S$ into a few pieces according to the sequence lengths to avoid too much padding that depletes disk storage space.

\paragraph{Parallel Verification.} During the generation process, \ppv selects the top $M$ candidate tokens $\vt = \{t_1,\cdots,t_M\} $ with the highest probabilities for each partially generated sequence $\va$.
Each $t_m$ is concatenated to $\va$, producing an array $\mathcal V$ for verification (\Cref{code:vector}). 
Then, simultaneously for each row $V_m$, \ppv performs a binary search on the sorted array $\mathcal X$ to identify the index $i_m$, where the row $\mathcal X_{i_m}$ is the smallest row that is lexicographically larger than the $V_m$. Since $A$ is ordered from left to right, \ppv can find the target in $\log(|\mathcal S|)$ time.
Importantly, we can derive that if $t_m \in \mathcal A_{\va, \mathcal S}$ if and only if $V_m \preceq \mathcal X_{i_m}$.
Therefore, a simple check tells whether $t_m$ is a valid subsequent token. 
Notice that all computations are performed on GPUs, and additional batching across multiple queries can further extend parallel processing.


\paragraph{Selection of top candidate tokens.} 
In \Cref{code:topk}, \ppv performs a sorting and selects only the top $M$ candidate tokens. 
In theory, we should set $M = |\mathcal V|$ to avoid missing valid tokens with low probability. 
However, we argue that this does not lead to a practical issue. 
First, most modern LLMs, when doing inference, will limit the choices to the tokens with top $50$ probabilities in the first place \citep{dubey2024llama}. 
This implies that verification essentially does not miss any valid candidates. In addition, our efficient parallel verification makes the verification cost almost identical when setting $M$ to $50$ or even $500$, as verified in the appendix. Additionally, the top $50$ probabilities already sum close to $1$, so there is no actual difference in setting $M$ to different values. 



\paragraph{Advantages over prefix trees.}
\ppv eliminates the need for CPU-based data structures like prefix trees as well as frequent data transfer between devices. In addition, its inference can be further parallelized across samples in a batch, leading to additional acceleration. Lastly, we notice that trie (often stored as JSON \citep{pezoa2016foundations}) loading is extremely slower compared with array loading. The evaluation time for a dataset can be $8.5$ times different at most.

%% file: Sections/experiments.tex
\input{Sections/large_table}

\section{EXPERIMENTS}

This section conducts comprehensive experiments to demonstrate the efficacy of our methods. Without further explanation, \alg is always equipped with \ppv.

\subsection{Experimental Settings}


\paragraph{Tasks \& datasets.} We evaluate on four tasks. The first two tasks are \emph{Entity Disambiguation}, which tries to find the most relevant entities from a list given an article, and \emph{Document Retrieval}, which tries to retrieve the most relevant documents. 
Each task contains six and eleven datasets, respectively.
\cite{Cao2021} fine-tunes BART models to generate entities using trie-based constrained decoding.
We follow their settings and generate entities from the set $\mathcal S$ that contains approximately six million Wikipedia pages. These datasets help compare our methods with the conventional trie-based approach using models fine-tuned on relevant tasks. Notice that \cite{Cao2021} provides results with additional tricks, such as manually setting $\mathcal S$ to a minimal candidate set for each query to achieve the best performance. Our paper seeks to provide a better constrained decoding algorithm, and it suffices to compare algorithms without using these tricks. That said, our method can be seamlessly combined with these add-ons. For instance, we demonstrate in the appendix that our method is still superior when combined with beam search.

Apart from existing benchmarks, we innovatively study the in-context learning setting for generative retrieval tasks, including \emph{Entity Retrieval} from DBPedia~\citep{hasibi2017dbpedia} and \emph{Fact Checking} with Wikipedia. We use the newly released LLAMA3.2-3B-INSTRUCT\footnote{\url{https://www.llama.com/}} as a generator. For each task, we select three query-answer pairs from the validation set and provide them in-context to the model to clarify the intentions and output formats. 
All datasets can be found from the information retrieval benchmark BEIR~\citep{thakur2021beir}.

\paragraph{Methods.} We compare vanilla generation (without constraints), trie-based constrained decoding, and \alg equipped with \ppv. For our method, the number of maximum sampling step $K$ is set to $\{1,2,3,4\}$, and the top candidate's size $M$ is set to $50$. 
Notice that when $K=1$, the performance of \alg should be the same as trie-based method (but faster). 
For sampling, we set model temperature $T$ to $1$ and do not include length penalty (rescale probabilities by length). We use A100 GPUs and set batch size to $128$. 

\paragraph{Evaluation metrics.}
For Entity Disambiguation and Document Retrieval, we adopt the \emph{InKB} micro-F1 following~\citet{le2018improving}, and R-precision~\citep{fischer2017privacy} following~\citet{petroni2020kilt} respectively. For in-context learning tasks, we utilize relevance score, which is calculated as the average query-document relevance using TREC evaluation tool~\citep{van2018pytrec_eval}.

\subsection{Main Results}

\begin{figure*}[t]
    \centering
    \includegraphics[width=1.0\textwidth]{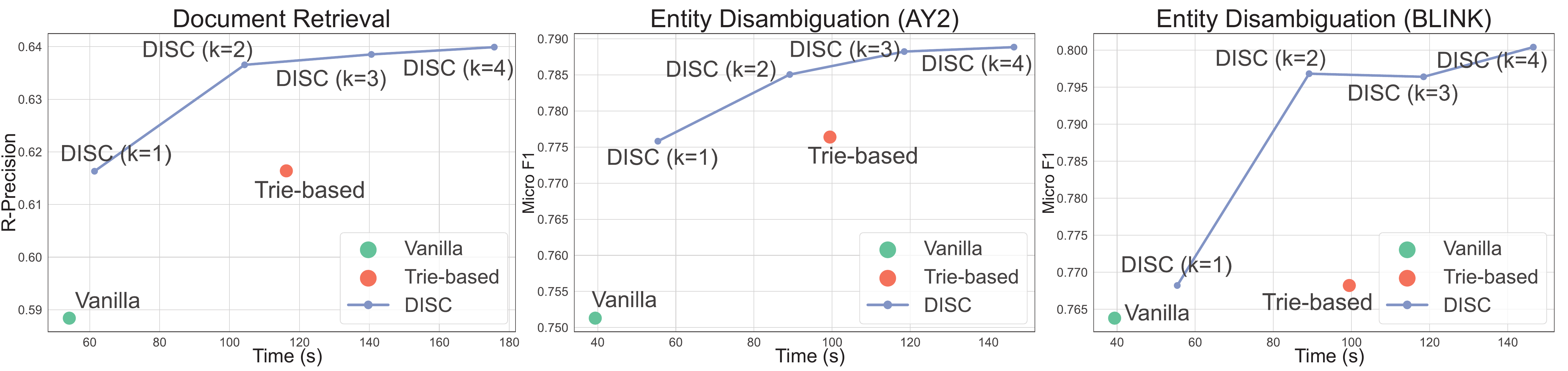}
    \vspace{-15pt}
    \caption{Time vs. Performance Comparison on Document Retrieval and Entity Disambiguation Tasks for all methods. The y-axis presents performance (R-Precision for Document Retrieval and Micro F1 scores for Entity Disambiguation), and the x-axis presents inference time that is average across all tasks belonging to that task.
    The improvement on time (x-axis) is resulted from \ppv, and the improvement on performance (y-axis) is resulted from \alg, respectively.
    }
    \label{fig:time-performance-curve}
    \vspace{-5pt}
\end{figure*}


\paragraph{Entity Disambiguation and Document Retrieval.} \Cref{tab:entity,tab:document} present the performance of unconstrained decoding (Vanilla), trie-based constrained decoding, and our method \alg with varying values of maximum sampling steps $K$. On both tasks, \alg consistently outperforms the trie-based constrained decoding across all datasets and models. On Entity Disambiguation with $K=2$, the average Micro F1 gain compared to Vanilla is increased from $0.1\%$ ($2.5\%$) to $3.3\%$ ($3.4\%$) for the BLINK (AY2) model.
For Document Retrieval, \alg achieves an average R-Precision gain of $5.2\%$, nearly double the $2.8\%$ improvement achieved by the trie-based method.
Remarkably, the effect of improving $K$ quickly plateaus when $K$ increases above $2$, while the difference between $K=1,2$ is non-trivial. 
The improvements are particularly pronounced on datasets like HoPo and NQ, where the R-Precision improvements are $21.58\%$ and $16.89\%$ compared to the Vanilla model, respectively.


The inference time for each algorithm and model is presented in \Cref{fig:time-performance-curve}. Results are averaged across all datasets. The advantages of \alg{} are evident, with $K = 1$ being significantly faster and $K=2$ offering much better performance while maintaining higher efficiency. The consistency of this optimal trade-off across all tasks clearly highlights the benefits of \ppv{}. 

\begin{table}[h]
\centering
\caption{The relevance score and inference time of all methods on Entity Retrieval and Fack Checking. Models are not tuned and only prompted in context.}
\label{tab:icl}
\resizebox{0.9\linewidth}{!}{\begin{tabular}{ccccc} 
\toprule
           & \multicolumn{2}{c}{\textbf{DBPedia-Entity}} & \multicolumn{2}{c}{\textbf{Climate-Fever}}  \\ 
\cmidrule{2-5}
           & Score                     & Time  (s)         & Score                    & Time      (s)       \\ 
\midrule
Vanilla    & 0.340                     & 20.20           & 0.160                    & 73.4             \\ 
\midrule
Trie-based & 0.600                     & 171.73          & 0.271                    & 234.8            \\ 
\midrule
\alg ($K=1$)     & 0.640                     & 20.41           & 0.270                    & 74.0             \\
\alg ($K=2$)     & 0.683                     & 37.92           & 0.294                    & 149.9            \\
\alg ($K=3$)    & 0.728                     & 58.04           & 0.290                    & 225.1            \\
\alg ($K=4$)     & \textbf{0.735}            & 76.88           & \textbf{0.295}           & 304.7            \\ 
\midrule
Delta (\%)      & \textcolor{red}{+116.18} &     -            & \textcolor{red}{+84.44} &                -  \\
\bottomrule
\end{tabular}}
\vspace{-1em}
\end{table}

\paragraph{Entity Retrieval and Fact Checking.} \Cref{tab:icl} presents the performance and efficiency comparison among different methods on the Entity Retrieval (DBPedia-Entity) and Fact Checking (Climate-FEVER) tasks.
On the first task, \alg with $K=4$ achieves a score of $0.735$, more than doubling the performance of Vanilla ($0.340$) and $22\%$ higher than the trie-based method.
Surprisingly, this is realized with nearly half of the inference time (since $|\mathcal S|$ is huge is this dataset).
Similarly, on the Climate-FEVER dataset, \alg with $K=4$ attains a score of $0.295$, marking an $84.4\%$ improvement over the Vanilla model and a $2.4\%$ increase over the trie-based method. 
As this in-context learning pipeline is novel and purely zero-shot, it clearly demonstrates the potential of \alg's capability to enhance performance without task-specific fine-tuning.









%% file: Sections/large_table.tex
\begin{table*}[t]
\centering
\caption{The Micro F1 score of unconstrained decoding (Vanilla), trie-based constrained decoding, and \alg on Entity Disambiguation across six datasets and two models. \textbf{Avg.} is the average score across datasets. The best score is bold, and $\Delta$ (in percentage) is the score improvement from Vanilla to \alg ($K=4$).}
\label{tab:entity}
\resizebox{\textwidth}{!}{\begin{tabular}{ccccccccccccccc} 
\toprule
\textbf{Dataset} & \multicolumn{2}{c}{\textbf{ACE2004}}                & \multicolumn{2}{c}{\textbf{AIDA}}                   & \multicolumn{2}{c}{\textbf{AQUAINT}}                & \multicolumn{2}{c}{\textbf{CWEB}}                   & \multicolumn{2}{c}{\textbf{MSNBC}}                  & \multicolumn{2}{c}{\textbf{WIKI}}                   & \multicolumn{2}{c}{\textbf{Avg.}}                  \\
\midrule
\textbf{Models}                & BLINK                    & AY2                      & BLINK                    & AY2                      & BLINK                    & AY2                      & BLINK                    & AY2                      & BLINK                    & AY2                      & BLINK                    & AY2                      & BLINK                    & AY2      \\
\midrule
Vanilla         & 0.794                   & 0.802                   & 0.830                   & 0.770                   & 0.791                   & 0.792                   & 0.630                   & 0.615                   & 0.761                   & 0.742                   & 0.778                   & 0.787                   & 0.764                   & 0.751                 \\
\midrule
Trie            & 0.809                   & 0.848                   & 0.830                   & 0.796                   & 0.792                   & 0.816                   & 0.629                   & 0.635                   & 0.770                   & 0.761                   & 0.779                   & 0.803                   & 0.768                   & 0.776               \\
\midrule
\alg ($K=1$)          & 0.809                   & 0.841                   & 0.830                   & 0.797                   & 0.792                   & 0.814                   & 0.629                   & 0.635                   & 0.770                   & 0.765                   & 0.779                   & 0.803                   & 0.768                   & 0.776                \\
\alg ($K=2$)          & 0.844                   & 0.848                   & 0.852                   & 0.803                   & 0.813                   & 0.821                   & 0.664                   & 0.647                   & 0.803                   & 0.781                   & 0.805                   & 0.811                   & 0.797                   & 0.785                \\
\alg ($K=3$)          & 0.844                   & 0.848                   & 0.852                   & 0.808                   & \textbf{0.813}          & \textbf{0.831}          & 0.665                   & \textbf{0.651}          & 0.796                   & \textbf{0.782}          & 0.808                   & 0.810                   & 0.796                   & 0.788                \\
\alg ($K=4$)          & \textbf{0.844}          & \textbf{0.849}          & \textbf{0.860}          & \textbf{0.810}          & 0.812                   & 0.827                   & \textbf{0.667}          & 0.651                   & \textbf{0.810}          & 0.781                   & \textbf{0.810}          & \textbf{0.816}          & \textbf{0.800}                   & \textbf{0.789}    \\
\midrule
$\Delta$ (\%)          & \textcolor{red}{+6.37} & \textcolor{red}{+5.89} & \textcolor{red}{+3.63} & \textcolor{red}{+5.27} & \textcolor{red}{+2.61} & \textcolor{red}{+4.34} & \textcolor{red}{+5.95} & \textcolor{red}{+5.80} & \textcolor{red}{+6.41} & \textcolor{red}{+5.13} & \textcolor{red}{+4.13} & \textcolor{red}{+3.73} & \textcolor{red}{+4.79} & \textcolor{red}{+5.00} \\

\bottomrule
\end{tabular}}
\vspace{-5pt}
\end{table*}

\begin{table*}[t]
\centering
\caption{The R-Precision of all methods on Document Retrieval across eleven datasets. Similar to \Cref{tab:entity}.}
\label{tab:document}
\resizebox{\textwidth}{!}{
\begin{tabular}{ccccccccccccc} 
\toprule
\textbf{Dataset} & \textbf{AY2}            & \textbf{CWEB}           & \textbf{ELI5}            & \textbf{FEV}            & \textbf{HoPo}            & \textbf{NQ}              & \textbf{zsRE}           & \textbf{TREx}            & \textbf{TQA}             & \textbf{WNED}           & \textbf{WoW}            & \textbf{Avg.}        \\
\midrule
Vanilla         & 0.884                   & 0.647                   & 0.0982                  & 0.757                   & 0.237                   & 0.484                   & 0.823                   & 0.675                   & 0.544                   & 0.818                   & 0.506                   & 0.588               \\
\midrule
Trie-based            & 0.894                   & 0.670                   & 0.104                   & 0.786                   & 0.265                   & 0.538                   & 0.860                   & 0.754                   & 0.556                   & 0.852                   & 0.502                   & 0.616               \\
\midrule
\alg ($K=1$)          & 0.900                   & 0.671                   & 0.104                   & 0.790                   & 0.266                   & 0.522                   & 0.862                   & 0.749                   & 0.561                   & 0.847                   & 0.508                   & 0.616                \\
\alg ($K=2$)          & 0.905                   & 0.683                   & 0.116                   & 0.803                   & 0.284                   & 0.563                   & 0.895                   & 0.773                   & 0.591                   & \textbf{0.863}         & 0.528                   & 0.637              \\
\alg ($K=3$)       & 0.907                   & 0.683                   & 0.113                   & 0.803                   & 0.287                   & 0.562                   & \textbf{0.906}         & \textbf{0.774}          & 0.601                   & 0.857                   & 0.531                   & 0.639               \\
\alg ($K=4$)        & \textbf{0.912}         & \textbf{0.687}         & \textbf{0.114}          & \textbf{0.804}         & \textbf{0.288}          & \textbf{0.566}          & 0.895                   & 0.773                   & \textbf{0.605}          & 0.863                   & \textbf{0.532}         & \textbf{0.640}      \\
\midrule
$\Delta$ (\%)           & \textcolor{red}{+3.17} & \textcolor{red}{+6.30} & \textcolor{red}{+15.5} & \textcolor{red}{+6.20} & \textcolor{red}{+21.6} & \textcolor{red}{+16.9} & \textcolor{red}{+8.81} & \textcolor{red}{+14.7} & \textcolor{red}{+11.2} & \textcolor{red}{+5.47} & \textcolor{red}{+5.11} & \textcolor{red}{+8.75} \\

\bottomrule
\end{tabular}}
\vspace{-5pt}
\end{table*}

%% file: Sections/related_works.tex
\section{RELATED WORKS }

\paragraph{Constrained decoding.} The objective of constrained decoding is to ensure that the generated sequence complies with the specified set of constraints. These constraints could be in the form of lexical constraints \citep{Hokamp2017, Anderson2017, Post2018} or regular expressions (REGEX) or grammatical structures defined by context-free grammars (CFGs) \citep{Geng2023, Kellner2024} or just a list of possible outputs \citep{Cao2021}.
Several approaches have been proposed to impose these constraints during the sequence generation process. For example, \citet{Cao2021} uses a trie structure, and \citet{Miao2019} utilizes Metropolis-Hastings sampling. While these approaches address the need to generate valid outputs that satisfy all the specified constraints, they often struggle with computational efficiency when the constraint set is large \citep{Stahlberg2019}. Therefore, generating valid outputs on-the-fly without exhaustive enumeration remains a challenge.

\paragraph{Generative retrieval.}
Recently, new approaches for retrieval have emerged whereby LLMs directly generate relevant documents for a given query without explicitly computing embeddings or doing nearest neighbor search \citep{rajput24}. There are primarily two ways of doing this: (1) the entire set of documents is put in the context of LLM, and then LLM generates relevant documents from this context \citep{lee2024can}, and (2) documents are not put in the context but to make sure that the generated documents are from the retrieval set, the decoding process is constrained \citep{Cao2021}. In applications where the retrieval set is enormous, the first approach would require the LLM to have $\mathcal O(N)$ context length where $N$ is the size of documents. As of today, this is too computationally expensive. In this paper, we focus on the second set of approaches.


%% file: supplement.tex

%
%





%

%

\onecolumn
\aistatstitle{Efficient and Asymptotically Unbiased\\Constrained Decoding for Large Language Models\\(Supplementary Materials)}


\input{Appendices/proofs}

\input{Appendices/code}

\input{Appendices/addition_exp_settings}

\input{Appendices/add_results}


%% file: Appendices/proofs.tex
\section{Proofs of Theorems}
\label{app:proof}

\subsection{Proof of \Cref{thm:alg}}
The logic of the proof can be decomposed into two parts. First, we compute the probability that upon $K$ rounds of samples, at least one sample is accepted. We demonstrate that the distribution will be unbiased in this case. Second, when $K$ new samples are re-drawn, we demonstrate that the importance sampling can upper bound the remaining KL divergence. Specifically, assume we use $Q$ to denote the sequence distribution of \Cref{code:IS}, i.e., the distribution that $K$ samples $\{\va_k\}_{k=1}^K$ are drawn using constrained decoding and further resample based on scores $\{x_k\}_{k=1}^K$. Then we upper bound $KL(P_\mathcal S \| Q)$ and use it to derive the eventual bound.

We first derive a few quantities useful for the proof.
According to \Cref{alg:disc}, $x$ is the importance score depending solely on sequence $\va$, and we therefore rewrite it as $$x(\va) = \prod_{i \in \{1, \cdots, |\va|\}} P_L( \mathcal A_{\va_{<i}, \mathcal S} | \va_{<i}).$$Intuitively, $x(\va)$ denotes the probability of alternatives in $\mathcal S$ along the token paths, which is smaller than the probability of all alternatives (which should be $1$ at each decoding step).
This is because at each decoding step, due to the application of constrained decoding, only tokens in $\mathcal A_{\va_{<i}, \mathcal S}$ are considered, leading to a total candidate probability smaller than 1. 
Readers are encouraged to verify that $\sum_{\va \in \mathcal S } P_L(\va) / x(\va) = 1$. 

With this, the distribution for one sampled sequence using constrained (denote as $\hat P$) is 
\begin{align*}
    \hat P(\va) = \frac {P_L(\va) / x(\va)} {\sum_{\va \in \mathcal S } P_L(\va) / x(\va)} = \frac {P_L(\va) }{ x(\va)}.
\end{align*}

\paragraph{Probability of acceptance.}
Consider one particular draw $\va_k$ among the $K$ rounds. The acceptance of $\va_k$ is $x(\va_k)$, and in expectation, each draw is accepted with probability 
$$
\sum_{\va \in \mathcal S} \hat P(\va) x(\va) = P_L(\mathcal S) = 1 - p_b.
$$
Therefore, at least one sample is accepted during $K$ rounds of sample are $1 - p_b^K$.


\paragraph{Unbiased distribution for the accepted sample.} To show that when a sample is accepted during $K$ rounds, we only need to show that for each round, the distribution is unbiased. Conditioned on acceptance, the probability that $\va \in \mathcal S$ is returned is $\hat P(\va) x(\va)/ (1-p_b) = P_L(\va) / (1-p_b) = P_\mathcal S (\va)  $.

\paragraph{KL Divergence between $P_\mathcal S$ and $Q$.} With probability $p_b^K$, $K$ rounds of samples are all rejected, and \alg has to resample $K$ samples and conduct importance sampling. 
We aim to analyze the KL divergence between the target distribution $P_\mathcal{S}$ and the distribution $Q$ resulting from importance sampling. 

Since the buffer size is $K$, the probability $Q(\va)$ is $K$ times the probability that sequence $\va$ is in the first position (which is $\hat P(\va)$) and subsequently wins during the importance sampling step (due to symmetry among the $K$ sequences). 
The challenge lies in computing the winning probability $W(\va)$ of sequence $\va$.

Specifically, we have 
\begin{align*} W(\va) = \sum_{\va_2, \ldots, \va_K \in \mathcal{S}} \prod_{j=2}^K \hat P(\va_j) \cdot \frac{x(\va)}{\sum_{j=1}^K x(\va_j)}, 
\end{align*} 
where $x(\va)$ is the importance score defined previously: 
\begin{align*} 
x(\va) = \prod_{i=1}^{|\va|} P_L(\mathcal{A}_{\va{<i}, \mathcal{S}} | \va_{<i}). 
\end{align*}

The KL divergence between $P_\mathcal{S}$ and $Q$ can then be expressed as (for simplicity, we set $A = 1 - p_b$)
\begin{align*} 
\mathrm{KL}(P_\mathcal{S} \| Q) 
&= \sum_{\va \in \mathcal{S}} \frac{P_L(\va)}{A} \ln \frac{P_L(\va) / A}{Q(\va)}  \\
&= \sum_{\va \in \mathcal{S}} -\frac{P_L(\va)}{A} \ln \frac{A K P_L(\va) W(\va)}{P_L(\va) x(\va)} \\
&= \sum_{\va_1 \in \mathcal{S}} -\frac{P_L(\va_1)}{A} \ln \left( \sum_{\va_2, \ldots, \va_K \in \mathcal{S}} \prod_{j=2}^K \hat P(\va_j) \cdot \frac{A K}{\sum_{j=1}^K x(\va_j)} \right) \\
&= \sum_{\va_1 \in \mathcal{S}} -\frac{\hat P(\va_1) x(\va_1)}{A} \ln \left( \sum_{\va_2, \ldots, \va_K \in \mathcal{S}} \prod_{j=2}^K \hat P(\va_j) \cdot \frac{A K}{\sum_{j=1}^K x(\va_j)} \right) .
\end{align*}
Applying Jensen's inequality (noting that $\ln \mathbb{E}[f] \geq \mathbb{E}[\ln f]$ for a convex function $\ln$), and considering the negative sign, we obtain 
\begin{align*} 
\mathrm{KL}(P_\mathcal{S} \| Q) &\leq \sum_{\va_1 \in \mathcal{S}} -\frac{\hat P(\va_1) x(\va_1)}{A} \sum_{\va_2, \ldots, \va_K \in \mathcal{S}} \prod_{j=2}^K \hat P(\va_j) \ln \left( \frac{A K}{\sum_{j=1}^K x(\va_j)} \right) \\
&= \sum_{\va_1, \ldots, \va_K \in \mathcal{S}} \prod_{j=1}^K \hat P(\va_j) \left( -\frac{x(\va_1)}{A} \right) \ln \left( \frac{A K}{\sum_{j=1}^K x(\va_j)} \right) \\
&= \sum_{\va_1, \ldots, \va_K \in \mathcal{S}} \prod_{j=1}^K \hat P(\va_j) \cdot \frac{x(\va_1)}{A} \ln \left( \frac{\sum_{j=1}^K x(\va_j)}{A K} \right). 
\end{align*}

Due to the symmetry among the indices $\{1,\cdots, K\}$ in the summation (each $\va_j$ is drawn independently and identically), we can replace $x(\va_1)$ with any $x(\va_j)$ without affecting the overall value. Therefore, we can generalize the expression by summing over all positions $j$ and averaging: 
\begin{align*} 
\mathrm{KL}(P_\mathcal{S} \| Q) 
&\leq \frac{1}{K} \sum_{r=1}^K \sum_{\va_1, \ldots, \va_K \in \mathcal{S}} \prod_{j=1}^K \hat P(\va_j) \cdot \frac {x(\va_r)}{A} \ln \left( \frac{\sum_{j=1}^K x(\va_j)}{A K} \right) \\
&=\sum_{\va_1, \ldots, \va_K \in \mathcal{S}} \prod_{j=1}^K \hat P(\va_j) \cdot \frac {\sum _{j=1}^K x(\va_j)}{AK} \ln \left( \frac{\sum_{j=1}^K x(\va_j)}{A K} \right) .
\end{align*}
Imagine $ \tilde X = (\va_1,\cdots, \va_K)$ is a random variable equaling to $ \frac{\sum_{j=1}^K x(\va_j)}{AK}$ with probability $\tilde P(\tilde X) = \prod_{j=1}^K \hat P(\va_j)$. Then, we have
\begin{align*}
    KL(P_\mathcal S \| Q) &\leq \sum \tilde P(\tilde X) f(\tilde X) = \mathbb E_{\tilde X} f(\tilde X),
\end{align*}
where $f(x) \triangleq x \ln x \leq |x-1|+ \frac{(x-1)^2}{2} $. This implies that
\begin{align}
    KL(P_\mathcal S \| Q) &= \mathbb E_{\tilde X} [|\tilde X -1|] + \mathbb E_{\tilde X}[\frac{(\tilde X - 1)^2}{2}] \\
    &\leq \sqrt{\mathbb E_{\tilde X}[(\tilde X -1)^2]} + \frac 1 2 \mathbb E_{\tilde X}[(\tilde X -1)^2], \label{eq:KL_upper}
\end{align}
where the inequality is by Jensen's inequality. The problem is therefore converted to upper bounding the variance of $\tilde X$ (whose expectation is $1$ because the expectation of $x(\va)$ is $A$). Due to the independence of each $\va_j$, we have
\begin{align*}
    \text{Var}(\tilde X) = \frac 1 K \text{Var}(\frac{\va}{A}) = \frac 1 K \text{Var}(\frac{\va }{A}) \leq \frac{A(1-A)}{KA^2},
\end{align*}
where the inequality is because all r.v. with a value ranging from $[0,1] $ and expectation $\xi$ has a variance upper bound by $\xi (1-\xi)$. Plugging this into the \Cref{eq:KL_upper}, we have
\begin{align*}
    KL(P_\mathcal S \| Q)  \leq \sqrt{\frac{p_b}{K(1-p_b)}} + \frac {p_b}{2K(1-p_b)}.
\end{align*}


\paragraph{KL Divergence between $P_\mathcal S$ and $P_{IS}^K$.} The final step is to bound the divergence of \alg using all the above results. According to the acceptance probability and the unbiasedness of accepted samples, we have 
\begin{align*}
    P^K_{IS} = p_b^K Q + (1-p_b^K) P_\mathcal S.
\end{align*}
The final bound is obtained by using \Cref{lemma}.

\begin{lemma}\label{lemma}
For any two distribution $P,Q$, any scalar $t \in [0,1]$, we have
\begin{align*}
     KL(P\|tP + (1-t)Q) \leq (1-t) KL(P\|Q).
\end{align*}
\end{lemma}

\subsection{Proof of \Cref{thm:lower_bound}}
To rigorously prove the theorem, we only need to construct an LLM output distribution, determine the candidate set $\mathcal S$, and derive the divergence. In fact, assume there are only two tokens, i.e., $\mathcal V = \{v_1, v_2\}$, and all sequences are with length $2$ (so four possible sequences in total, with an ending $\texttt{EOK}$ token after each sequence being draw with probability $1$. We assume that 
$$
P_L(v_1v_1) = P_L(v_1v_2) = \frac{1-p_b}2, P_L(v_2v_1 )=p_b\epsilon, P_L(v_2v_2) = p_b(1-\epsilon).
$$

Assume $\mathcal S = \{v_1v_1,v_1v_2,v_2v_1\}$. In this case, the ground-truth distribution $P_{\mathcal S} $ is 
$$
P_L(v_1v_1) = P_L(v_1v_2) = \frac 1 2 \cdot \frac{1 - p_b}{1 - p_b + p_b\epsilon} , P_L(v_2v_1 )=\frac{p_b \epsilon}{1 - p_b + p_b\epsilon} .
$$
On the contrary, if we apply constrained decoding, then for the first token to be decoded, both $v_1, v_2$ are valid. Therefore, the probability of choosing $v_1, v_2$ is $1-p_b$ and $p_b$, separately. When $v_1$ is chosen, as $v_1v_1$ and $v_1v_2$ belong to $\mathcal S$, the probability for each of them conditioned on $v_1$ being chosen is $\frac 1 2$. However, when $v_2$ is chosen, since $v_2v_2$ does not satisfy the constraint, the model can only choose $v_2v_1$. This gives the distribution of $P_\mathcal S^{CD}$ as 
$$
P_\mathcal S^{CD} (v_1 v_1) = P_\mathcal S^{CD}(v_1v_2) =  \frac {1-p_b} 2, P_\mathcal S^{CD}(v_2v_1) = p_b.
$$

Consequently, the KL divergence between the two distributions is (denote $\frac{1 - p_b}{1 - p_b + p_b\epsilon} $ as $\kappa$
\begin{align*}
    KL(P_\mathcal S \| P_\mathcal S ^{CD}) &= 2\times \frac 1 2 \kappa \ln \frac{\kappa / 2}{(1-p_b) / 2} + \frac{p_b \epsilon}{1 - p_b + p_b \epsilon}\ln (\frac{p_b \epsilon}{p_b(1 - p_b + p_b \epsilon ) }  ) \\
    &= \kappa \ln \frac{\kappa}{1 - p_b} + \frac{p_b \epsilon}{1 - p_b + p_b \epsilon}\ln (\frac{ \epsilon}{(1 - p_b + p_b \epsilon ) }  ).
\end{align*}
The final result is obtained by considering $\epsilon \to 0$.

%% file: Appendices/code.tex
\section{\texttt{Torch} code for \ppv}

\lstset{style=mypython}
\begin{lstlisting}[caption={Parallel Prefix-Verification (PPV) algorithm}]
def ppv(P, a, X, M, vocab_size):
    """
    Args:
        P (torch.Tensor): Token distribution tensor of shape (vocab_size,).
        a (torch.Tensor): Partial sequence tensor of shape (len_a,).
        X (torch.Tensor): Constraint matrix of shape (N, l_max), sorted lexicographically.
        M (int): Number of top tokens to consider.
        vocab_size (int): Size of the vocabulary.
    Returns:
        torch.Tensor: Mask of valid tokens of shape (vocab_size,).
    """
    # Get top M token indices
    _, vt = torch.topk(P, M)  # vt: (M,)

    # Build candidate sequences V by appending vt to a
    V = torch.cat([a.unsqueeze(0).repeat(M, 1), vt.unsqueeze(1)], dim=1).to(X.device)  # V: (M, len_a + 1)

    # Initialize binary search bounds
    N = X.shape[0]
    low = torch.zeros(M, dtype=torch.long, device=X.device)
    high = torch.full_like(low, N)

    # Perform binary search
    while (low < high).any():
        mid = torch.div(low + high, 2, rounding_mode="floor")
        mid_seqs = X[mid][:, :V.shape[1]]  # Shape: (M, len_a + 1)

        # Lexicographical comparison
        cmp = (V > mid_seqs).float() - (V < mid_seqs).float()
        cmp_result = cmp.cumsum(dim=1)[:, -1]  # Final comparison result per sequence

        # Update bounds based on comparison results
        greater = cmp_result > 0
        low = torch.where(greater, mid + 1, low)
        high = torch.where(~greater, mid, high)

    # Check for matches after binary search
    valid = (low < N)
    V_valid = V[valid]
    matching_seqs = X[low[valid]][:, :V.shape[1]]

    # Verify exact matches
    matches = (V_valid == matching_seqs).all(dim=1)
    valid_indices = vt[valid][matches]

    # Create mask of valid tokens
    mask = torch.zeros(vocab_size, dtype=torch.bool, device=X.device)
    mask[valid_indices] = True
    return mask
\end{lstlisting}

%% file: Appendices/addition_exp_settings.tex
\newpage

\section{Additional Experimental Settings}

\paragraph{Entity Disambiguation.} We use two BART checkpoints provided by~\citet{Cao2021} to retrieve entities from $\sim$6M Wikipedia pages. Two of the BART checkpoints are used to evaluate on Entity Disambiguation:
The first BART checkpoint, denoted as \texttt{BART-BLINK}, is pre-trained on the BLINK data~\citep{wu2019scalable}, i.e., 9M unique document-entity pairs from Wikipedia. The other BART checkpoint, \texttt{BART-AIDA} is further fine-tuned on AIDA-CoNLL dataset~\citep{hoffart2011robust}. We test Entity Disambiguation with 6 datasets: AIDA-CoNLL, MSNBC, AQUAINT, ACE2004, WNED-CWEB (CWEB) and WNED-WIKI (WIKI)~\citep{guo2018robust}.

\paragraph{Document Retrieval.} We use the BART checkpoint for Document Retrieval provided by~\citet{Cao2021}, which is trained on all the BLINK and KILT data~\citep{petroni2020kilt}. Document Retrieval is tested with KILT benchmark~\citep{petroni2020kilt}, consisting fact checking with FEVER~\citep{thorne2018fever}; open-domain question answering using Natural Questions~\citep{adelani2021masakhaner}, HotpotQA~\citep{yang2018hotpotqa}, TriviaQA~\citep{joshi2017triviaqa}, ELI5~\citep{fan2019eli5}; slot filling with T-REx~\citep{elsahar2018t}, Zero Shot RE~\citep{levy2017zero}; entity disambiguation on AIDA CoNLL-YAGO, WNED-WIKI and WNED-CEWB; dialog with Wizard of Wikipedia~\citep{dinan2018wizard}. 

\paragraph{Entity Retrieval \& Fact checking} Without fine-tuning, the language model is provided with a few demonstrations from the target corpus and leverages its in-context learning ability to generate entities accordingly in these two tasks.  For each task, three query-doc pairs from the validation set are provided as in-context examples to demonstrate the task intentions and output formats. We use the datasets from zero-shot information retrieval benchmark BEIR~\citep{thakur2021beir}, i.e., DBPedia-Entity~\citep{hasibi2017dbpedia}, Climate-FEVER~\citep{diggelmann2020climate} for Entity Retrieval and Fact Checking respectively. The prompts are given as follows:

\begin{itemize}
    \item \begin{verbatim}
Retrieve entity from DBpedia for my query.
Query: <what is foreign policy for angola>    Entity: <Foreign relations of Angola>
Query: <animalia>    Entity: <Animalia (book)>
Query: <what is asphalt and what is it used for>    Entity: <Asphalt>
Query: <{$newQuery}>    Entity:
\end{verbatim}
    \item \begin{verbatim}
Retrieve entity from Wikipedia for my query.
Query: <Global warming is driving polar bears toward extinction>    
Entity: <Habitat destruction>
Query: <The sun has gone into ‘lockdown’ which could cause freezing weather, earthquakes
and famine, say scientists>    
Entity: <Famine>
Query: <the Great Barrier Reef is in fine fettle>    
Entity: <Great Barrier Reef>
Query: <{$newQuery}>    
Entity:
\end{verbatim}
\end{itemize}

%% file: Appendices/add_results.tex
\section{Additional Results}

\begin{table}[h]
\centering
\caption{The Micro F1 score and evaluation time for \texttt{DISC} with different $K$ (allowed sampling steps) and $M$ (top candidate tokens). The task is Entity Disambiguation across six datasets with BART AY2 checkpoint. \textbf{Avg.} is the average score across datasets. }
\resizebox{\textwidth}{!}{
\begin{tabular}{ccccccccccccc!{\vrule width \lightrulewidth}cc} 
\toprule
\multirow{2}{*}{} & \multicolumn{2}{c}{\textbf{ACE2004}} & \multicolumn{2}{c}{\textbf{AIDA}} & \multicolumn{2}{c}{\textbf{AQUAINT}} & \multicolumn{2}{c}{\textbf{CWEB}} & \multicolumn{2}{c}{\textbf{MSNBC}} & \multicolumn{2}{c!{\vrule width \lightrulewidth}}{\textbf{WIKI}} & \multicolumn{2}{c}{\textbf{Avg.}}  \\
                  & F1     & Time                        & F1     & Time                     & F1     & Time                        & F1     & Time                     & F1     & Time                      & F1     & Time                                                    & F1     & Time                      \\ 
\midrule
\multicolumn{15}{c}{$K=1$}                                                                                                                                                                                                                                                                                             \\ 
\midrule
$M=1000$            & 0.8405 & 3.377                       & 0.8379 & 55.436                   & 0.7799 & 9.089                       & 0.6484 & 142.65                   & 0.7927 & 8.946                     & 0.8009 & 89.52                                                   & 0.7834 & 51.50                     \\
$M=500$             & 0.8483 & 2.961                       & 0.8395 & 52.747                   & 0.7951 & 8.636                       & 0.6504 & 135.50                   & 0.7668 & 8.350                     & 0.8003 & 86.47                                                   & 0.7834 & 49.11                     \\
$M=100$             & 0.8483 & 2.730                       & 0.8459 & 49.429                   & 0.7937 & 8.104                       & 0.6518 & 127.27                   & 0.7774 & 8.553                     & 0.7948 & 81.23                                                   & 0.7853 & 46.22                     \\
$M=50$              & 0.8327 & 2.736                       & 0.8421 & 48.054                   & 0.7992 & 7.912                       & 0.6517 & 126.86                   & 0.7835 & 8.121                     & 0.7965 & 80.15                                                   & 0.7843 & 45.64                     \\ 
\midrule
\multicolumn{15}{c}{$K=2$}                                                                                                                                                                                                                                                                                             \\ 
\midrule
$M=1000$            & 0.8521 & 4.751                       & 0.8535 & 86.152                   & 0.8157 & 13.944                      & 0.6616 & 226.15                   & 0.7973 & 13.358                    & 0.8038 & 147.77                                                  & 0.7973 & 82.02                     \\
$M=500$             & 0.8483 & 4.558                       & 0.8531 & 83.668                   & 0.8088 & 14.043                      & 0.6641 & 212.14                   & 0.7851 & 12.798                    & 0.8047 & 139.36                                                  & 0.7940 & 77.76                     \\
$M=100$             & 0.8405 & 4.641                       & 0.8546 & 80.060                   & 0.8198 & 12.915                      & 0.6634 & 204.21                   & 0.7896 & 12.761                    & 0.8046 & 133.91                                                  & 0.7954 & 74.75                     \\
$M=50$              & 0.8560 & 4.224                       & 0.8493 & 74.577                   & 0.8061 & 12.575                      & 0.6618 & 195.74                   & 0.8003 & 11.783                    & 0.8043 & 126.26                                                  & 0.7963 & 70.86                     \\ 
\midrule
\multicolumn{15}{c}{$K=3$}                                                                                                                                                                                                                                                                                             \\ 
\midrule
$M=1000$            & 0.8249 & 7.056                       & 0.8544 & 123.357                  & 0.8143 & 22.207                      & 0.6652 & 337.48                   & 0.8049 & 18.632                    & 0.8102 & 217.88                                                  & 0.7956 & 121.10                    \\
$M=500$             & 0.8327 & 6.166                       & 0.8562 & 109.944                  & 0.8226 & 18.060                      & 0.6623 & 281.26                   & 0.7988 & 16.643                    & 0.8102 & 187.58                                                  & 0.7971 & 103.27                    \\
$M=100$             & 0.8405 & 5.447                       & 0.8493 & 98.833                   & 0.8171 & 16.148                      & 0.6642 & 256.41                   & 0.7973 & 15.214                    & 0.8097 & 168.73                                                  & 0.7963 & 93.46                     \\
$M=50$              & 0.8405 & 5.878                       & 0.8542 & 102.032                  & 0.8157 & 17.204                      & 0.6646 & 263.40                   & 0.8018 & 15.205                    & 0.8059 & 171.65                                                  & 0.7971 & 95.89                     \\ 
\midrule
\multicolumn{15}{c}{$K=4$}                                                                                                                                                                                                                                                                                             \\ 
\midrule
$M=1000$            & 0.8405 & 7.806                       & 0.8549 & 137.484                  & 0.8322 & 22.944                      & 0.6668 & 357.17                   & 0.8018 & 20.691                    & 0.8081 & 241.67                                                  & 0.8007 & 131.29                    \\
$M=500$             & 0.8366 & 7.246                       & 0.8549 & 134.291                  & 0.8184 & 21.457                      & 0.6635 & 344.06                   & 0.7896 & 19.285                    & 0.8094 & 223.37                                                  & 0.7954 & 124.95                    \\
$M=100$             & 0.8444 & 7.065                       & 0.8555 & 130.161                  & 0.8171 & 21.470                      & 0.6655 & 331.44                   & 0.8018 & 19.462                    & 0.8080 & 221.01                                                  & 0.7987 & 121.77                    \\
$M=50$              & 0.8327 & 6.680                       & 0.8598 & 121.303                  & 0.8088 & 20.674                      & 0.6667 & 318.97                   & 0.8079 & 18.179                    & 0.8069 & 211.05                                                  & 0.7971 & 116.14                    \\
\bottomrule
\end{tabular}}
\label{tab:M}
\end{table}

\begin{table}[h]
\centering
\caption{The Micro F1 score for \texttt{DISC} with different beam search sizes with different $K$. The task is Entity Disambiguation across six datasets with BART AY2 checkpoint. \textbf{Avg.} is the average score across datasets. We show that \texttt{DISC} can be combined with beam search decoding strategy to improve performance.}
\resizebox{\textwidth}{!}{
\begin{tabular}{ccccccccccccc|cc} 
\toprule
\textbf{Dataset} & \multicolumn{2}{c}{\textbf{ACE2004}} & \multicolumn{2}{c}{\textbf{AIDA}} & \multicolumn{2}{c}{\textbf{AQUAINT}} & \multicolumn{2}{c}{\textbf{CWEB}} & \multicolumn{2}{c}{\textbf{MSNBC}} & \multicolumn{2}{c|}{\textbf{WIKI}} & \multicolumn{2}{c}{\textbf{Avg.}}  \\ 
\midrule
Beam size        & 1 & 2                      & 1 & 2                   & 1 & 2                      & 1 & 2                   & 1 & 2                    & 1 & 2                    & 1 & 2                    \\ 
\midrule
\texttt{DISC} ($K=1$)           & 0.841 & 0.852                     & 0.797 & 0.869                   & 0.814 & 0.824                     & 0.635 & 0.673                  & 0.765 & 0.803                   & 0.803 & 0.818                   & 0.776 & 0.807                    \\
\texttt{DISC} ($K=2$)           & 0.848 & 0.856                     & 0.803 & 0.871                  & 0.821 & 0.827                     & 0.647 & 0.676                  & 0.781 & 0.806                    & 0.811 & 0.819                   & 0.785 & 0.809                    \\
\texttt{DISC} ($K=3$)           & 0.848 & 0.860                     & 0.808 & 0.872                  & 0.831 & 0.825                     & 0.651 & 0.677                  & 0.782 & 0.797                   & 0.810 & 0.820                   & 0.788 & 0.809                    \\
\texttt{DISC} ($K=4$)           & 0.849 & 0.860                     & 0.810 & 0.873                  & 0.827 & 0.825                     & 0.651 & 0.678                  & 0.781 & 0.805                   & 0.816 & 0.820                   & 0.789 & 0.810                    \\
\bottomrule
\end{tabular}}
\label{tab:beam}
\end{table}

In this section, we present additional experimental results to show that \texttt{DISC} doesn't require huge $M$ (the number of top candidate tokens) and is compatible with other advanced decoding strategies such as beam search. In~\cref{tab:M}, we evaluate \texttt{DISC} on Entity Disambiguation task with $M$ ranges from $\{50, 100, 500, 1000\}$. From the results in \Cref{tab:M}, we observe that reducing the value of $M$ from $1000$ to $50$ has a minimal impact on the Micro F1 scores across all datasets. For instance, with $K=1$, decreasing $M$ from $1000$ to $50$ only results in a slight fluctuation in F1 scores—often within a margin of $0.01$. This indicates that even a small set of top candidate tokens ($M=50$) is sufficient for \texttt{DISC} to achieve near-optimal performance.
Moreover, a smaller $M$ significantly reduces the evaluation time. For example, with $K=1$ on the CWEB dataset, the evaluation time drops from $142.65$ seconds at $M=1000$ to $126.86$ seconds at $M=50$. This demonstrates that choosing a smaller $M$ not only maintains performance but also enhances computational efficiency.
Increasing the allowed sampling steps $K$ generally improves the F1 scores across datasets, as seen when comparing results from $K=1$ to $K=4$. However, this comes at the cost of increased evaluation time. For example, with $M=1000$, the average time increases from $51.50$ seconds at $K=1$ to $131.29$ seconds at $K=4$. Therefore, there’s a trade-off between performance gains and computational overhead when adjusting $K$.

In \Cref{tab:beam}, we evaluate the compatibility of \texttt{DISC} with beam search decoding strategies. Incorporating beam search with a beam size of $2$ consistently improves the Micro F1 scores across all datasets and values of $K$. For instance, at $K=1$, the average F1 score increases from $0.776$ with greedy decoding to $0.807$ with beam search. This enhancement suggests that beam search helps explore more candidate sequences, leading to better disambiguation outcomes.
Combining beam search with higher values of $K$ continues to yield marginal improvements. However, the gains taper off as $K$ increases, indicating that most of the performance benefits are captured with smaller $K$ and beam sizes. This synergy between \texttt{DISC} and beam search showcases the method’s flexibility and its ability to integrate with advanced decoding strategies to further boost performance.

In summary, these additional experiments demonstrate that \texttt{DISC} is both efficient and adaptable. It does not require a large number of top candidate tokens to perform effectively, and it can seamlessly incorporate beam search to enhance its disambiguation capabilities. This makes \texttt{DISC} a practical choice for entity disambiguation tasks where computational resources and time are critical considerations.